# An alternative for one-hot encoding in neural network models

Lazar Zlatić,

zlaticlazar@gmail.com

## Abstract

This paper proposes an algorithm that implements binary encoding of the categorical features of neural network model input data, while also implementing changes in the forward and backpropagation procedures in order to achieve the property of having model weight changes, that result from the neural network learning process for certain data instances of some feature category, only affect the forward pass calculations for input data instances of that same feature category, as it is in the case of utilising one-hot encoding for categorical features.

## 1. Introduction

In order to successfully utilise neural network models on input data containing one or more categorical features, such models must implement some sort of encoding of the said categorical data into numerical values in order for that data to be used in the neural network learning process. In general there are several different types of encoding that can be utilised, offering different advantages and disadvantages depending on the available data and the model structure [2,3].

One of the ways to encode categorical data is to use one-hot encoding, which replaces the categorical feature with a set of features whose cardinality is equal to that of the set of possible feature categories, such that, for a certain instance of input data, all of the feature values are set to zero, except for the one that is numbered the same as the category of the considered data instance (for some arbitrary chosen numbering order), which is set to one. One-hot encoding offers the advantage of considering the contributions of input data instance belonging to each category, to the output separately, by encoding the categories in mutually orthogonal vectors. However one-hot encoding is impractical for features that take categories from a set of large cardinality.

Another way to encode categorical features is to use binary encoding which implements the replacement of the original categorical feature with a number of features on the order of the $log_2$ of the cardinality of the set of the categories, in such a way that the value of each of the new features is either one or zero corresponding to the bitwise representation of the number assigned to each category (again in an arbitrary numbering order of categories). Such an approach provides an advantage over one-hot encoding for high cardinality sets of categories. However, it does not have the property of the mutual independence of the features that one-hot encoding has. For example, a change in the weight assigned to a connection between a neuron representing one of the binary encoding features and a neuron of the following layer will affect all the forward pass calculations for instances of data that have a feature category number whose bitwise representation shares a one on the some position.

Following section of this paper proposes an addition to the standard forward and backpropagation procedures [1], in order to achieve the property of models implementing one-hot encoding of having the change of the model weights due to the backpropagation process for an input data instance of a certain feature category only affect the forward pass calculations for data instances of the same feature category, while using fewer encoding features (neurons) as in the binary encoding.

## 2. Decription of the proposed alternative procedure

Consider first applying one-hot encoding on a categorical feature of input data in the case of using a neural network machine learning model. Let $C$, $|C| = N$, be the set of all categories that can be assigned to a feature of an instance of the input dataset. For now consider only the case where each dataset instance can only be attributed one category for the given feature. In the case of a neural network model, this means that the first (input) layer of the neural network will have $N$ neurons (among others) whose values encode the assigned feature category in such a way that the value of each neuron is $x_i = \delta_{ic}$, where $i = 1,2,\ldots,N$; $c$ is the number, given in some chosen order, of the attribute category for the considered instance of data, $c \in \{1,2,\ldots N\}$; and $\delta_{ic} = \begin{cases} 1, & i = c \\ 0, & i \neq c \end{cases}$.

Let $K$ be the number of neurons in the layer following the input layer connected to the considered $N$ encoding neurons such that there is a connection between any one of these $K$ neurons and the considered $N$ encoding neurons of the input layer. The value $O_k$, for $k = 1,2,\ldots,K,$ assigned to each of these $K$ neurons is calculated as:

$$O_k = F\left(\sum_{i=1}^{N} w_{ik}x_i + \sum_{i \in others} w_{ik}x_i + b_k\right) \quad (1)$$

where $F$ is the chosen activation function, and the second sum in the expression goes over all other neurons in the input layer apart from the $N$ encoding neurons. Weight $w_{ik}$ is the weight assigned to the connection between the neuron $i$ of the input layer, and the neuron $k$ of the following layer, $x_i$ indicates the value of the neuron $i$ of the input layer, while $b_k$ is the bias of the neuron $k$ of the following layer.

In such a model, performing backpropagation, after previously doing forward propagation for a certain instance of data assigned with category number $c$ of the considered feature, will result in the change of only those weights $w_{ck}$, for $k = 1,2,\ldots,K$, that are attributed to the connections of the neuron corresponding to the feature category of the input data instance. Thus, these weights are changed such that their new values are:

$$w_{ck}^{new} = w_{ck} - \varepsilon grad(O_k)F' \quad (2)$$

where $\varepsilon$ is the used learning rate for the model, $grad(O_k)$ is the previously calculated (in the process of backpropagation) gradient corresponding to the $k$th neuron of the following layer, and $F'$ is the derivative of the chosen activation function with respect to its input.

Consider now the alternative neural network structure where the $N$ encoding neurons of the input layer are replaced with $n = \lceil log_2(N + 1) \rceil$ neurons (the $\lceil \ \rceil$ is used to symbolise rounding up), all connected to each of the same $K$ neurons of the following layer. We want for the used model to have the same property, as the previously described model which utilises one-hot encoding, of having the changes of the connection weights resulting from the backpropagation process, following after the forward propagation for a certain instance of the input data with an assigned considered attribute $c$, only affect the values of the $K$ neurons of the following layer when the forward propagation calculations are performed for an instance of data with the same assigned category $c$. In order to achieve this, the calculations performed during the forward and backpropagation processes have to be altered.

Consider the following algorithm for the alternative neural network learning procedure:

1. Firstly, we form two matrices $A$ and $B$, of sizes $N \times K$ and $K \times n$ respectively, and initialise all their values to 0.

2. The forward propagation process is altered in such a way that, each time we perform forward propagation in order to calculate the values in the $K$ neurons of the second layer, we add $A_{ck}$ to and subtract all $B_{ki}, i \in BR(c)$, from $O_k$, where $A_{ck}$ is the element of matrix $A$ in the row numbered with the feature category number $c$ in the column $k$, while the counter $i$ indicating the columns of the elements of $B$, considered within the calculation, takes values from the set $BR(c)$ of the numbers of the positions of ones in the bitwise representation of $c$ (for example if $c = 1101$, the set $BR(c) = \{1,3,4\}$ ). In the matrix representation the previous calculation can be viewed as:

$$O_K = W \times X + BK + A[c] - B \times BRV[c] \quad (3)$$

where $O_K, W, X$ and $BK$ represent the vector of the values in the $K$ neurons of the following layer, the matrix of connection weights between the input and the following layer, the vector of input data feature values and the vector of bias values, respectively, $A[c]$ is the vector equal (in element values) to the row $c$ of $A$, while $BRV[c]$ is the vector corresponding to the bitwise representation of $c$ (for the previous example of $c = 1101$, $BRV[c] = [1\ 0\ 1\ 1]^T$ ). Here $A[c] - B \times BRV[c]$ has the role of an additional bias component to the calculations of $O_K$.

3. The backpropagation process is altered in such a way that, at the end of backpropagation after updating the values of connection weights according to the calculated gradients, we also perform an update of the elements of matrices $A$ and $B$ such that $A_{ck}^{new} = A_{ck} - \varepsilon grad(O_k)F'$ and $B_{ki}^{new} = B_{ki} - \varepsilon grad(O_k)F'$, for $i \in BR(c)$. Here $A$ and $B$ have the role of parameters being updated during the learning process.

All the other calculations within the forward and backpropagation processes, including the updating of the weight parameters, remain the same as in the first proposed regular neural network structure.

The proposed algorithm utilises the memorisation matrix $A$ to keep track of the change in weight values of encoding neurons corresponding to the backpropagation after a certain input instance with the considered feature of category $c$, and the memorisation matrix $B$ to keep track of the change in weight values of encoding neurons regardless of category $c$. Thus, using $A[c] - B \times BRV[c]$ as bias in forward propagation calculations for an instance of input data with the considered feature of category $c$, works as to ensure that only the change in weights that resulted from the backpropagation for the instances of data with the considered feature of category $c$ contribute to the value of $O_k$.

One can notice the redundant changes of weight parameters $w_{ik}, i \in BR(c)$, and matrix elements $B_{ki}, i \in BR(c)$, that are later subtracted during forward pass calculations. This was done in order to make it so that the proposed algorithm differs from the standard learning procedure only in the addition of calculations involving $A$ and $B$ matrices whose values are stored and updated separately from the rest of the learning procedure.

While in the standard one-hot encoding structure we perform updates on $N \times K$ weights corresponding to encoding neurons, in the case of the proposed structure we perform updates on encoding neuron weights and $A$ and $B$ matrix elements in the order of $n \times K$ operations, each backpropagation pass. Similarly, in each forward pass, for each $k = 1, 2, \ldots, K$, there are in the order of $n$ operations, corresponding to the encoding neuron weights and $A$ and $B$ matrix elements, for calculating $O_k$, as opposed to in the order of $N$ operations in the standard one-hot encoding model. However, since the vectors representing the feature category, formed by one-hot encoding, are sparse, calculations involving them are in general faster than calculations of non-sparse vectors of the same size.

As element $A[c] - B \times BRV[c]$ in expression (3) can also be viewed as an additional bias in forward pass calculations, one can also implement the algorithm retaining expression (3) in the usual form:

$$O_K = W \times X + BK \qquad (4)$$

while adjusting $BK$ at the begining of each forward pass by adding $A[c] - B \times BRV[c]$ to its value.

One can also consider the possibility of an input data instance being assigned multiple categories for a considered feature. This case can be considered as a separate category and the same proposed procedure can be performed, with diminishing returns, in terms of possible benefits compared to standard one-hot encoding, as the number of ways to assign multiple categories for a given feature increases.

Also, in the case of having a different structure, in terms of connections between the encoding and neurons of the following layer (not the fully connected case presented here), the proposed algorithm has to be adapted, if possible, to the required structure.

## 3. Conclusion

The proposed algorithm can provide an advantage to the standard one-hot encoding procedure in terms of the number of performed operations (during the learning process) and the size of the neural network structure, while retaining the advantageous property of one-hot encoding in terms of separating connection weight changes resulting from backpropagation for different feature category input data.

However, the model still has the same space complexity due to the formation of matrix $A$. Also, one must consider alterations to the proposed algorithm when considering neural networks of different structures in terms of connections between the input and the following layer.

## References

[1] Rumelhart, David E., Geoffrey E. Hinton, and Ronald J. Williams. "Learning representations by back-propagating errors." *nature* 323.6088 (1986): 533-536.

[2] Seger, Cedric. "An investigation of categorical variable encoding techniques in machine learning: binary versus one-hot and feature hashing." (2018).

[3] Potdar, Kedar, Taher S. Pardawala, and Chinmay D. Pai. "A comparative study of categorical variable encoding techniques for neural network classifiers." *International journal of computer applications* 175.4 (2017): 7-9.